\documentclass[letterpaper, 10 pt, conference]{ieeeconf}  

\IEEEoverridecommandlockouts                              

\usepackage{multicol}
\usepackage[bookmarks=true]{hyperref}
\usepackage{xcolor}
\usepackage{amsmath}
\usepackage{graphicx}
\usepackage{grffile}
\usepackage{blindtext}
\usepackage[bottom]{footmisc}
\usepackage{mathrsfs}
\usepackage{lipsum}
\usepackage{url}
\usepackage{array}
\usepackage{booktabs}
\usepackage{float}

\newlength\fwidth
\usepackage[linesnumbered,ruled]{algorithm2e}
\usepackage{graphicx}
\usepackage{framed}
\usepackage{subcaption}
\usepackage{indentfirst}
\usepackage{multirow}
\usepackage{makecell}

\usepackage{amsmath, amssymb,amsthm}
\usepackage{booktabs}
\usepackage{tabularx}
\usepackage[capitalise]{cleveref}
\usepackage{comment}
\usepackage{romannum}
\usepackage{color}
\let\labelindent\relax
\usepackage{enumitem}

\title{\LARGE \bf
Learning Dense Rewards for Contact-Rich Manipulation Tasks
}

\author{Zheng Wu$^{1,2}$, Wenzhao Lian$^{1}$, Vaibhav Unhelkar$^{3}$, Masayoshi Tomizuka$^{2}$ and Stefan Schaal$^{1}$
\thanks{$^{1}$X, the Moonshot Factory, Mountain View, CA 94043 USA. $^{2}$University of California, Berkeley, Berkeley, CA 94704 USA. $^{3}$Rice University, Houston, TX 77005 USA. This research was conducted during Zheng's internship at X.}%
}

\begin{document}

\maketitle
\thispagestyle{empty}
\pagestyle{empty}

\begin{abstract}
Rewards play a crucial role in reinforcement learning. To arrive at the desired policy, the design of a suitable reward function often requires significant domain expertise as well as trial-and-error. Here, we aim to minimize the effort involved in designing reward functions for contact-rich manipulation tasks. In particular, we provide an approach capable of extracting dense reward functions algorithmically from robots’ high-dimensional observations, such as images and tactile feedback. In contrast to state-of-the-art high-dimensional reward learning methodologies, our approach does not leverage adversarial training, and is thus less prone to the associated training instabilities. Instead, our approach learns rewards by estimating task progress in a self-supervised manner. We demonstrate the effectiveness and efficiency of our approach on two contact-rich manipulation tasks, namely, peg-in-hole and USB insertion. The experimental results indicate that the policies trained with the learned reward function achieves better performance and faster convergence compared to the baselines.
\end{abstract}

\IEEEpeerreviewmaketitle

\section{Introduction}
\label{sec:intro}

Reinforcement learning (RL) has shown promising performance on a variety of complex tasks, ranging from playing video games to vision-based robotic control \cite{sutton2018reinforcement, mnih2015human, kalashnikov2018scalable}. However, the success of RL techniques relies heavily on the availability of high-quality, dense reward signals. Learning policies on tasks with sparse rewards, such as Montezuma’s revenge \cite{mnih2015human}, remains notoriously challenging. In robotics, these dense rewards often need to be hand-crafted by a human. This \emph{reward engineering} typically requires significant domain expertise as well as trial-and-error. In contact-rich manipulation tasks, due to the discontinuous dynamics and high-dimensional observation spaces, the challenge of reward specification is further amplified; thus, making the adoption of RL methods for robotic manipulation difficult in practice.

The goal of our work is to reduce the effort involved in reward specification for contact-rich manipulation tasks, such as peg-in-hole and connector insertion~\cite{kimble2020benchmarking}. Inverse reinforcement learning (IRL) techniques, which learn reward functions from expert demonstrations, offer one such alternative to circumvent the challenge of reward engineering \cite{abbeel2004apprenticeship}. However, classical IRL techniques require hand-engineered states or features, making their application to tasks with high-dimensional and continuous observation spaces (such as camera images and tactile feedback) challenging \cite{abbeel2004apprenticeship, ramachandran2007bayesian, ziebart2008maximum}. More recently, by leveraging generative adversarial learning \cite{goodfellow2014generative}, IRL techniques have been developed to address continuous \cite{finn2016guided, fu2017learning}, high-dimensional observation spaces \cite{singh2019end, mees2020adversarial} and successfully demonstrated on robotic manipulation tasks. However, despite their encouraging performance, this class of methods inevitable inherits the instabilities associated with adversarial training \cite{salimans2016improved} and fail to utilize multi-modal observations. 

\begin{figure}
  \centering
  \includegraphics[width=\linewidth]{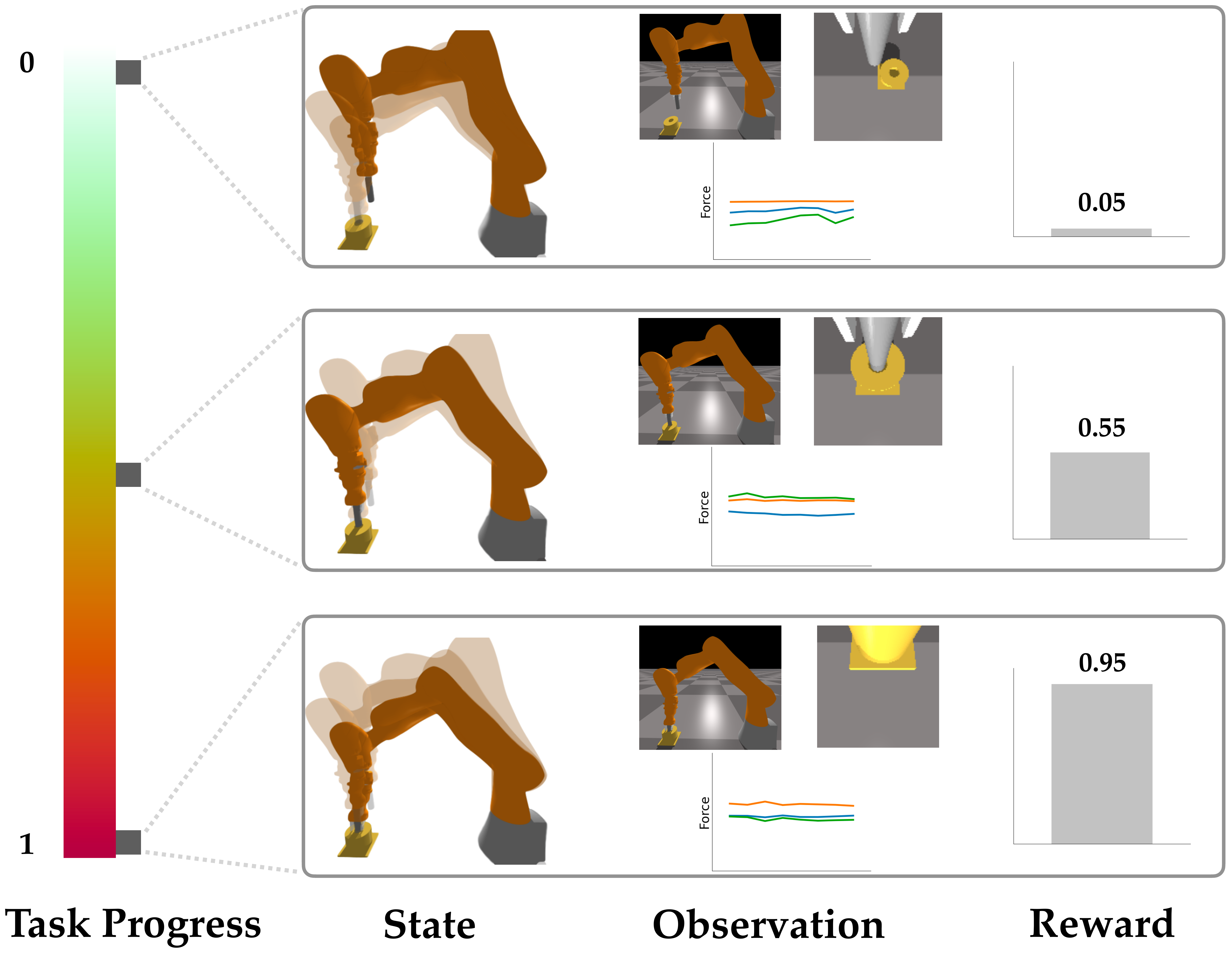}
  \caption{We provide a reward learning approach (\texttt{DREM}) based on the notion of \emph{task progress}. For instance, in the figure, the robot is performing a peg-in-hole task. As the robot moves closer to the peg, the learned reward signal increases, thereby facilitating sample efficient training of RL algorithms. \texttt{DREM} arrives at this reward function algorithmically by learning a mapping from robot's multi-modal observations to task progress in a self-supervised manner.}
  \vspace{-5mm}
  \label{fig:intro:insight}
\end{figure}

To learn reward functions for contact-rich manipulation tasks while avoiding the challenges of generative adversarial learning, we propose a novel approach \textbf{Dense Rewards for Multimodal Observations} \texttt{(DREM)}. Several contact-rich manipulations tasks can be specified through sparse reward signals, such as a goal state or a Boolean measure of task success (e.g., a peg being inside a hole, electricity passing through a connector and socket pair). While these readily available sparse reward signals are often insufficient for sample efficient high-dimensional RL, they can help design dense reward signals when coupled with expert demonstrations or, even, self-supervision. Our approach, thus, combines a sparse reward signal, self-supervision, and (optionally) observation-only expert demonstrations to extract a dense reward function. This dense reward can then be used with any appropriate RL technique to arrive at the robot policy.

Given the sparse measure of task success and optional expert demonstrations, \texttt{DREM} learns a mapping from the high-dimensional observation space to a latent measure of \emph{task progress}. For an arbitrary robot observation, this latent measure aims to provide its task-specific distance to the goal state, thereby serving as a proxy for the dense reward signal. Intuitively, our approach builds on the insight that a state that is closer to the task goal (in a task-specific metric) should be assigned a higher reward than one that is farther, as shown in Fig.~\ref{fig:intro:insight}. Our core contribution is to realize this insight computationally for high-dimensional, multimodal, and continuous observation spaces. \texttt{DREM} achieves this by first creating a dataset of observations and their corresponding task progress (latent space labels) through an efficient, self-supervised sampling process. This dataset is then used to train an encoder-decoder network, attain a latent space, and obtain the dense reward function.

We evaluate the proposed approach to reward learning on two representative contact-rich manipulation tasks: peg-in-hole and USB insertion.
In these evaluations, \texttt{DREM} learns the reward using only \textit{one} expert demonstration.
The learned reward function is then used to generate robot policies using Soft Actor-Critic, a model-free RL algorithm \cite{haarnoja2018soft}.
Experimental results show that the policies trained with our learned reward achieves better performance and faster convergence compared to baseline reward functions, including those obtained by recent reward learning approaches.

\section{Problem Statement}
\label{sec:problem}

\subsection{Tasks of Interest}
Motivated by near-term applications in assembly, we focus on contact-rich robotic manipulation tasks. In particular, we consider tasks that can be suitably modeled as discrete-time Markov decision processes (MDPs) \cite{puterman2014markov}. Briefly, MDPs describe sequential decision-making problems and are parameterized via the tuple $\mathcal{M} \equiv (\mathcal{S}, \mathcal{A}, \mathcal{T}, R)$, where $\mathcal{S}$ corresponds to the state space,  $\mathcal{A}$ corresponds to the action space, $\mathcal{T}$ denotes the transition model, and $R(s) \rightarrow \Re$ represents the real-valued reward of state $s$.


In contact-rich manipulation tasks, the state space is composed of high-dimensional, continuous observations available from robot sensors.
The robot observation is typically multi-modal, e.g., visual (through cameras), tactile (through force-torque sensors), and proprioceptive (through measurements of joint angles and velocities).
The action space models the actuation capability of the robot. Prior works have explored a variety of action spaces across the joint space, the operational space, and their combinations \cite{martin2019variable}. Our problem is agnostic to the specification of the action space. In our analysis, without loss of generality, we model actions as the end-effector twist (linear and angular velocity) of the robot.

The transition model represents the physics of the environment.
Due to the task being contact-rich, the model is highly non-linear, potentially discontinuous, and difficult to specify analytically. 
This motivates the need for model-free RL techniques.
The objective of an agent solving the MDP is to arrive at a policy $\pi(s) \rightarrow a$ that maximizes its expected cumulative reward.
Several contact-rich manipulation tasks encountered in the manufacturing domain can be suitably described by the above MDPs, such as peg-in-hole, connector insertion, and gear assembly \cite{kimble2020benchmarking}.

\subsection{Inputs and Outputs}
Assembly tasks are typically categorized by their \emph{repeatability} and presence of a \emph{goal} state $s_G$, which can be used to arrive at a sparse reward (i.e., a positive value at $s_G$ and zero otherwise).
We denote this sparse goal reward as $R_G$.
However, as motivated in Section~\ref{sec:intro}, this sparse signal is often inadequate for learning the policy for high-dimensional continuous MDPs, requiring prohibitive amounts of exploration.
Instead, to achieve sample efficiency, RL techiques require a richer reward signal that can guide exploration and result in faster convergence to the optimal policy.

Hence, our problem focuses on learning a dense reward function (denoted as $R$), which when provided as the reward signal to RL techniques results in the optimal policy for the MDP, 
$\mathcal{M} \equiv (\mathcal{S}, \mathcal{A}, \mathcal{T}, R_G)$.
To solve this problem, we assume access to the MDP state and action specifications, the sparse reward function $R_G$, a training environment (i.e., robot's real or simulated environment), and (optionally) expert observation-only demonstrations\footnote{Acquiring measurements of actions is challenging when collecting demonstrations from human experts \cite{argall2009survey}, motivating the focus on observation-only demonstrations.}, i.e., state sequences from an expert performing the task. 
Further, we delimit our scope to manipulation tasks where task progress is monotonic (i.e., the optimal policy does not induce cycles in the state space).
In relation to RL, the two rewards $R_G$ and $R$ can be viewed as extrinsic (specified by the human designer) and intrinsic (self-supervised) reward signals of the robot, respectively \cite{singh2009rewards}.

\section[Learning Rewards for Multimodal Observations]{Learning Rewards\\ for Multimodal Observations}
\label{sec:method}


\begin{figure*}[!htb]
  \centering
  \includegraphics[width=0.88\textwidth]{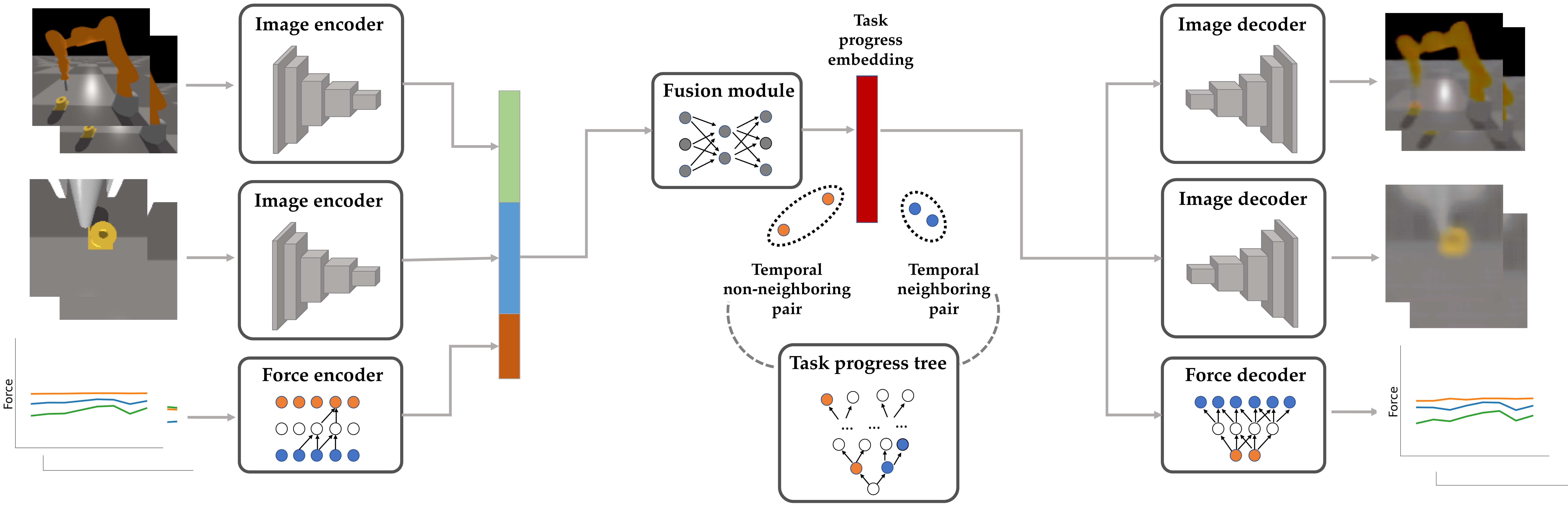}
  \caption{\texttt{DREM} takes three sensor signals: RGB images from a fixed camera, RGB images from a wrist-mounted camera, and 6-axis force data of the last $8$ frames from a F/T sensor. The latent embedding is learned using a reconstruction loss (defined between the input and the decoded output) and a triplet loss (defined on the temporal neighboring and non-neighboring pairs as detailed in Section~\ref{sec:method:representation_learning}, where the data pairs are sampled from the task progress tree in Section~\ref{sec:method:backward_sampling}).}
  \label{fig:method:model_overview}
  \vspace{-6mm}
\end{figure*}


The core idea of our method is to learn a continuous variable $p \in [0, 1]$ from the observation to represent the current progress towards completing the given task.
We name it the \textit{task progress} variable. 
When $p = 0$, it indicates the robot is at the initial state and $p = 1$ indicates the task is finished. 
If task progress is learned for each observation, it can then be directly used as the reward for policy learning algorithms, i.e., $R(s_t) = p(s_t)$.
We propose to learn the function $R(s)$, such that on a nominal task execution trajectory, the task progress variable is aligned with the temporal positions of the states observed.
Namely, for any nominal task execution trajectory $\tau = [s_0, s_1, ..., s_{N-1}]$ and $s_{N-1}$ coincides with the goal state $s_G$, we have
\begin{equation}
\label{eq:compare}
    t_2 - t_1 > \epsilon \quad \rightarrow \quad R(s_{t_2}) > R(s_{t_1})
\end{equation}
where $0 \leq t_1 < t_2 \leq N-1$ and $\epsilon > 0$.
Our approach uses this insight for learning an embedding from the observation space $\mathcal{S}$ to a latent space $\mathcal{H}$, $h_\phi: \mathcal{S} \rightarrow \mathcal{H}$, in which the task progress can be measured as the distance between any state $s_t$ and the goal state $s_G$. Once the embedding is learned, the reward function can be easily derived using the ratio of distance in the latent space:
\begin{equation}
\label{eq:reward_distance}
    R(s_t) = 1 - \frac{\text{dist}(h_\phi(s_t), h_\phi(s_G))}{\text{dist}(h_\phi(s_0), h_\phi(s_G))}.
\end{equation}

Our proposed framework, titled \texttt{DREM}, is composed of two stages. First, we adopt a novel backward sampling process, i.e., sampling from the goal state to the feasible start region, to create a dataset of observations $s$ and their temporal positions $p(s)$ . Second, we use the generated data to learn the embedding that satisfies the constraints defined in Eq.~\eqref{eq:compare} and Eq.~\eqref{eq:reward_distance}. It is worth noting that a \textit{single} observation-only expert trajectory is used in the backward sampling process and the embedding is learned in a self-supervised manner.

\subsection{Backward Sampling}
\label{sec:method:backward_sampling}
To learn the embedding $h_{\phi}$, state observations with different temporal positions are required as the training data. One straightforward solution is to collect many (tens to hundreds of) expert demonstrations and record the corresponding observations for each state. However, such data collection can be extremely time-consuming, making the adoption of our method difficult in practice. Thus, we propose a novel backward sampling process (i.e., sampling from the goal state to start region) to tackle the data generation problem. The similar idea was also explored in ~\cite{florensa2017reverse} for policy learning with increasing difficulty and ~\cite{zakka2020form2fit} for learning generalizable shape descriptor for kit assembly.

Our proposed sampling process is based on the insight that for most manipulation problems the variance of goal state is much smaller than the start state. For instance, consider the peg-in-hole task where goal state corresponds to the peg being inside the hole, while the start state corresponds to the robot holding the peg in free space. For this task, the goal state is  significantly more constrained than the start state (which has much larger variance), thereby making backward sampling more efficient compared to forward sampling. We demonstrate the efficacy of the backward sampling method with a single observation-only (without action annotations) expert demonstration available.

We achieve the backward sampling process by constructing a state tree, named as the \textit{task progress tree}, where each node represents a state observation and the node's depth represents its temporal position (in reverse order). A general framework to build task progress trees is detailed as follows:

\begin{enumerate}[leftmargin=0.5cm]
    \item Add goal state $s_G$ to an empty seed set $Q^0$ with maximum capacity $N$; assign the current depth $d$ to $0$.
    \item For each of the state $s_q$ in $Q^d$, sample $M$ random actions $\{a_k, k=1,2,...,M\}$ and apply each of them to $s_q$ to get the next state observations $\{s_{k|q}^{d+1}, k=1,2,...,M\}$. Add the $M$ next state observations into the tree as the child nodes of the parent node $s_q$.
    \item For each child node sampled in Step 2, compute an approximate progress measure $f(s_{k|q}^{d+1})$. Intuitively, the progress measure $f(\cdot)$ indicates whether the state observation is progressing away from the goal state. It can either be task-agnostic such as measuring the information gain on the distribution of visited states \cite{settles2009active}, or task-specific such as using one or multiple observation-only expert trajectories as a reference. The latter option is adopted in our implementation and detailed below.
    \item Select $N$ samples from the $M\cdot N$ child nodes in Step 2 as the new seed set $Q^{d+1}$ based on the progress measure in Step 3; increment $d$ to $d+1$.
    \item Repeat Steps 2-4 until the number of the states in $Q^d$ that are in the start region is above a threshold $N \cdot \delta$, where $\delta \in (0,1)$. For instance, for the peg-in-hole task, the start region is defined as the set of states with robot's end effector position being above a pre-defined height threshold.
\end{enumerate}

As mentioned above, a single observation-only expert trajectory is used to construct the task-specific progress measure in our implementation (Step 3). Specifically, given the expert trajectory $\xi = \{s_e^0, s_e^1, ..., s_e^{N-1}\}$, we simplify Steps 2-4 by setting $N=1$. At depth $d$, we assign $Q^d$ as a single element set, containing the state closest to $s_e^{N-d-1}$ (the $d-$th state in reversed order of the expert trajectory $\xi$) measured by robot's end-effector pose difference. The obtained task progress tree, consists of state observation nodes along with their depths, equivalently, temporal positions (in reversed order). These states and temporal positions are then used as training data to train the desired embedding $h_{\phi}$. 



\subsection{Multi-Modal Representation Learning}
\label{sec:method:representation_learning}
We aim to learn the embedding from multi-modal inputs that satisfy Eq.~\eqref{eq:reward_distance} using the data generated in Section~\ref{sec:method:backward_sampling}. Fig.~\ref{fig:method:model_overview} illustrates our representation learning model.

\subsubsection{Multi-Modality Encoder}
\label{sec:method:representation_learning:encoder}
Three sources of sensory data are used as the inputs of our multi-modality encoder: RGB images from a fixed camera, RGB images from a wrist-mounted camera, and force-torque (F/T) readings from a wrist-mounted F/T sensor. We adopt a similar encoder architecture as used in~\cite{lee2019making}. For 128x128x3 RGB images, we use a 6-layer Convolutional Neural Network, followed by a fully-connected layer to transform each image to a 64 dimensional vector. For F/T readings, we take the last 8 readings from the 6-axis F/T sensor as a 8x6 time-series and perform 4-layer causal convolution~\cite{oord2016wavenet} to transform the F/T readings into a 32 dimensional vector. The three vectors are concatenated and passed as input to a 2-layer Multi-Layer Perceptron  to produce the final 128 dimensional hidden vector.

\subsubsection{Multi-Modality Decoder}
\label{sec:method:representation_learning:decoder}
The decoder takes the fused hidden representation as input and tries to reconstruct the multi-modal sensor input. For reconstructing RGB images, we use 7-layer transposed convolution to upsample the hidden vector to the original 128x128x3 size. For F/T readings, we use 3-layer 1-d transpose convolution to convert the data back to 8x6 size. The loss function is defined between the original sensor input and reconstructed output. We use Sigmoid loss for RGB images and $L_2$ loss for F/T readings.

\subsubsection{Task Progress Latent Representation Learning}
\label{sec:method:representation_learning:latent_learning}
Our goal is to obtain a latent space, where the distance measure between any state latent representation $h_{\phi}(s_t)$ and goal state latent representation $h_{\phi}(s_G)$ reflects the current \textit{task progress} (reward), as shown in Eq~\eqref{eq:reward_distance}. The supervision signal comes only from the corresponding temporal positions of the states. This is achieved by enforcing the prior that for any state pair $(s_{t_1}, s_{t_2})$ , when $t_1 < t_2$, $s_{t_2}$ should be closer to the goal state $s_G$ than $s_{t_1}$ in terms of the distance measure in the latent space. Let $g(s_{t_1}, s_{t_2})$ be the distance gap between $(s_{t_1}, s_G)$ and $(s_{t_2}, s_G)$ in the latent space, i.e., $g(s_{t_1}, s_{t_2}) = dist(h_{\phi}(s_{t_1}), h_{\phi}(s_G))-dist(h_{\phi}(s_{t_2}), h_{\phi}(s_G))$. We train the desired latent representation by defining the triplet loss as:

\[
L_{\text{triplet}} = 
     \begin{cases}
        \max[0, \beta_1 (t_2-t_1) - g(s_{t_1}, s_{t_2})], \\
        \qquad \qquad \qquad \qquad \qquad \text{if} \epsilon < (t_2 - t_1); \text{and} \\
        \max[0, g(s_{t_2}, s_{t_1})] + \max[0, g(s_{t_1}, s_{t_2})-\beta_2], \\
        \qquad \qquad \qquad \qquad \qquad \text{if} \ 
        0 \leq (t_2 - t_1) \leq \epsilon,
     \end{cases}
\]

where $\epsilon, \beta_1, \beta_2$ are hyper-parameters in our model. We set $\epsilon=4$, $\beta_1 = 0.04$, $\beta_2 = 0.04$. $\epsilon$ is the temporal margin we set to distinguish temporally neighboring pairs from non-neighboring pairs. For temporally neighboring pairs, we enforce them to have similar distances w.r.t. the goal state in the latent space. For non-neighboring pairs, we enforce states whose temporal positions are closer to the goal state should also be closer to the goal state in terms of the distance measure in the latent space. The distance metric in the latent space is defined as:
\begin{equation}
    \text{dist}(h_{\phi}(s_t), h_{\phi}(s_G)) = 1- \langle {\overline{h}_{\phi}(s_t)}, \overline{h}_{\phi}(s_G) \rangle,
\end{equation}
where $\langle \cdot, \cdot \rangle$ denotes the dot product and $\overline{h}$ denotes $L_2$ normalization. Intuitively, the distance from any state to the goal state is represented by $1-cos(\theta)$, where $\theta$ denotes the angle between the two corresponding embedding vectors.\footnote{We explored other distance metrics, e.g., the Euclidean distance, using which our method also works. We empirically found policies trained with rewards using the chosen metric achieves the fastest convergence.}

The whole network is trained using a weighted sum of the reconstruction loss and triplet loss, $L = L_{\text{triplet}} + \alpha L_{\text{reconstruction}}$. We set $\alpha = 0.2$ during training. To prevent the data imbalance issue, we collect the same number of temporal neighboring pairs and non-neighboring pairs when sampling from the task progress tree. Once the embedding is trained, we obtain the reward function according to Eq.~\eqref{eq:reward_distance}.

\section{Experiments}
\label{sec:experiment}
\begin{figure}
	\begin{subfigure}[t]{0.22\textwidth}
		\centering
		\includegraphics[width=\textwidth]{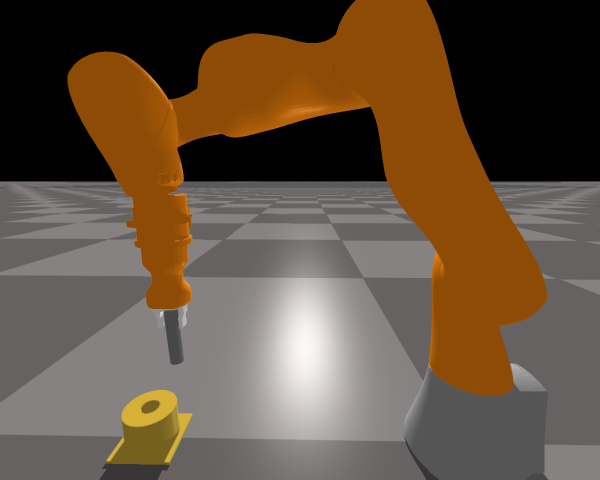}
		\caption{The peg-in-hole task}
	\end{subfigure}%
	~ 
	\begin{subfigure}[t]{0.22\textwidth}
		\centering
		\includegraphics[width=\textwidth] {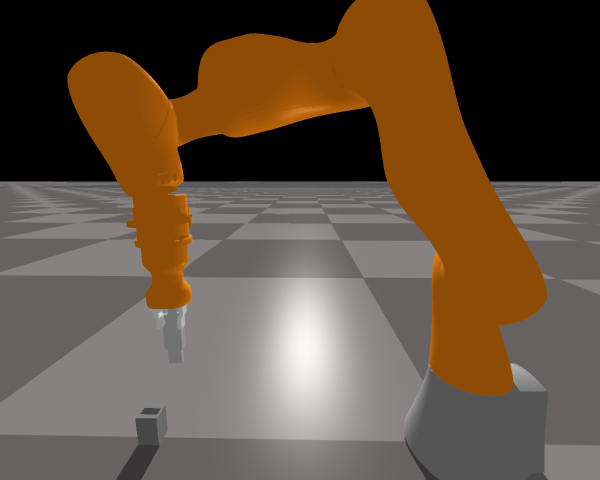}
		\caption{The USB insertion task}
	\end{subfigure}
	\caption{The two manipulation tasks used in our experiments.}
	\label{fig:experiment:tasks}
  \vspace{-5mm}
\end{figure}

We aim to investigate two questions in the experiments. First, we examine if the learned reward realizes our insight, namely, states that are closer to the task goal should have higher rewards. This allows us to evaluate the effectiveness of our learned reward. Second, we study how the learned reward can be applied to policy learning in acquiring contact-rich manipulation skills. Our study is based on the evaluation in two common contact-rich manipulation tasks: peg-in-hole and USB insertion (shown in Fig.~\ref{fig:experiment:tasks}).

\subsection{Experimental Setup}
All the experiments are conducted in Isaac Gym~\cite{liang2018gpu}, a GPU-accelerated simulator based on NVIDIA PhysX and Flex simulation backends, which provides high physics fidelity for contact-rich simulation. We defer validating our approach on real-robots to future work. The tasks are performed using a 7-DoF robotic arm modeled on the Kuka LBR IIWA robot. Three types of sensor observations are served as the input to the robot: 128x128 RGB images from a fixed camera, 128x128 RGB images from a wrist-mounted camera, and 6-axis F/T readings from a wrist-mounted F/T sensor. For the peg-in-hole task, we use a round peg with a clearance of 2.4mm. For the USB insertion task, the clearance is 1.0mm.
The simulated robot is torque controlled and the torque command is computed via the following control law:
\begin{equation}
\label{eq:control_law}
    \tau = - J^T K_D ( \Dot{x} - \Dot{x}_{ref}),
\end{equation}
where $\tau$ denotes the 7-dimension joint torque, $J$ the Jacobian, and $K_D$ the gain matrix. $\Dot{x}$ is the current end-effector twist, and $\Dot{x}_{ref}$ is the twist command (action) computed by an RL agent or a scripted policy. The joint torques are sent to the robot at 100Hz and the policies are queried at 5Hz.

\begin{figure}
  \centering
  \includegraphics[width=0.85\linewidth]{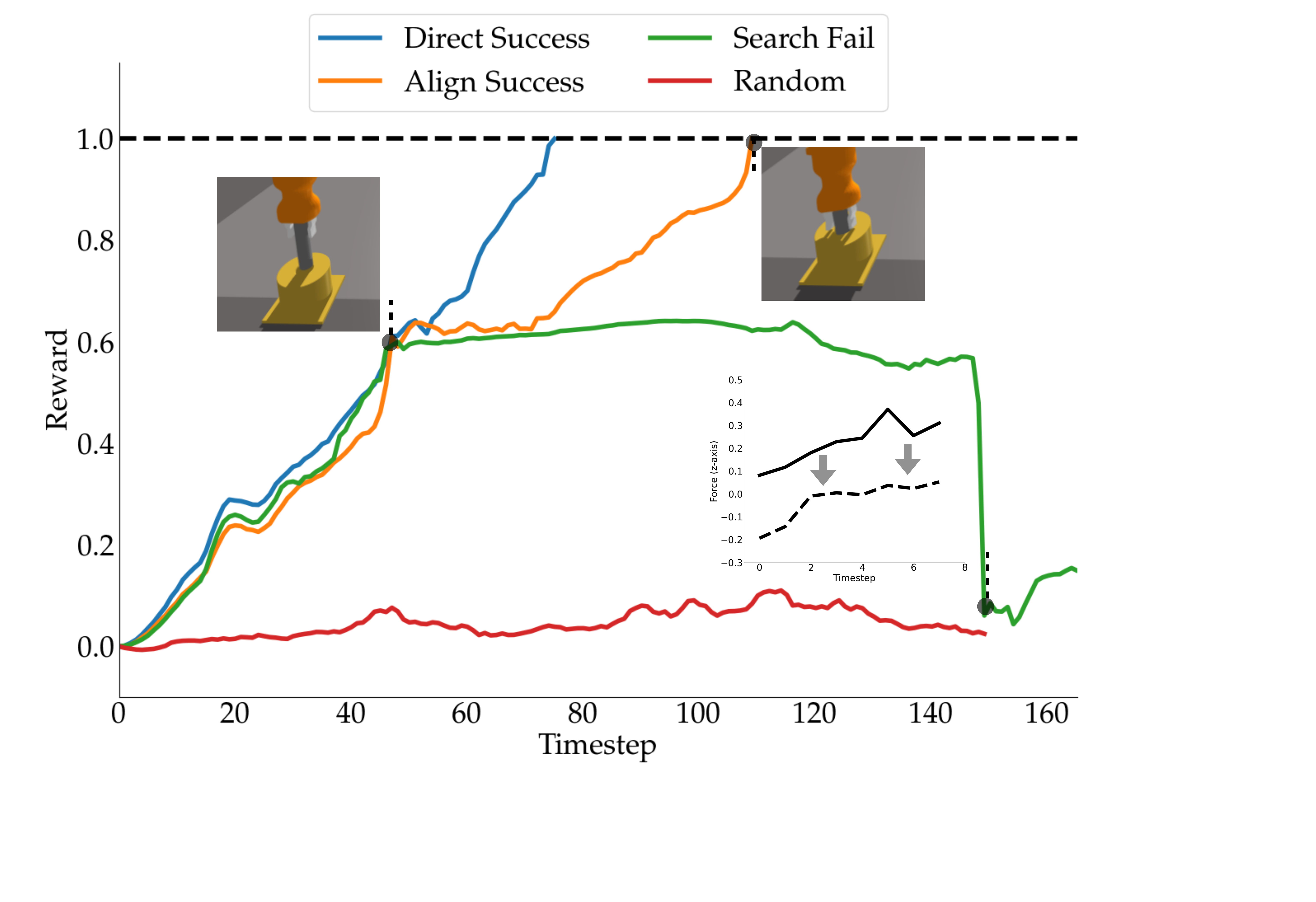}
  \caption{Visualization of the learned reward function on four different scripted policies.}
  \label{fig:experiment:reward_examin}
  \vspace{-7mm}
\end{figure}

\subsection{Implementation Details}
In the backward sampling process for both tasks, we set the number of random actions $M = 500$. We randomly perturb the colors of the objects (peg \& hole and USB male \& USB female) in the scene to prevent overfitting to certain colors. To train the latent embedding model, we adopted the network architecture detailed in Section~\ref{sec:method:representation_learning}. We trained the model on a Titan 2080Ti GPU for $20,000$ iterations using Adam~\cite{kingma2014adam} with learning rate $lr = 2e^{-4}$.

\subsection{Learned Reward Examination}
In this experiment, we want to examine the rationality of the learned reward. Specifically, we want to test if the output of our reward model assigns sensible values to the visited states under various policies. To achieve this, we test our learned reward model with four different scripted policies on the peg-in-hole task listed below.
\begin{enumerate}
    \item Imitate the expert demonstrator by directly inserting the peg into the hole (\texttt{Direct Success}).
    \item Approach the hole with a small translation error until contact $\rightarrow$ Align the peg with the hole center $\rightarrow$ Insert the peg into the hole (\texttt{Align Success}).
    \item Approach the hole with a small translation error until contact $\rightarrow$ Move away from the hole center while the peg is in contact with the hole $\rightarrow$ The peg loses contact with the hole (\texttt{Search Fail}).
    \item Random policy (\texttt{Random}).
\end{enumerate}
The results of the learned reward model on the four scripted policies are shown in Fig.~\ref{fig:experiment:reward_examin}. In \texttt{Direct Success}, the reward increases (approximately) linearly from $0$ to $1$ along with the number of timesteps in the trajectory. The reward in \texttt{Align Success} first increases similarly as in \texttt{Direct Success}, until the peg makes contact with the hole surface. Then it roughly stays the same while attempting to align with the hole center, and continues increasing to 1 when the peg is aligned and being inserted into the hole. For \texttt{Search Fail}, the reward behaves similarly as in \texttt{Align Success} at the beginning (increases and then plateaus). The reward then abruptly decreases to a small value (below $0.1$). That is exactly the timestep when the peg loses contact with the hole surface (the force in z-axis rapidly changes from a large positive value to around $0$). The reward of \texttt{Random} just fluctuates around a small value close to $0$.
The above observations demonstrate the interpretability of our learned reward model and its potential to provide denser supervision for policy learning.

\subsection{RL Policy Learning with Learned Reward}
\begin{figure}
    \centering
    \includegraphics[width=0.83\linewidth]{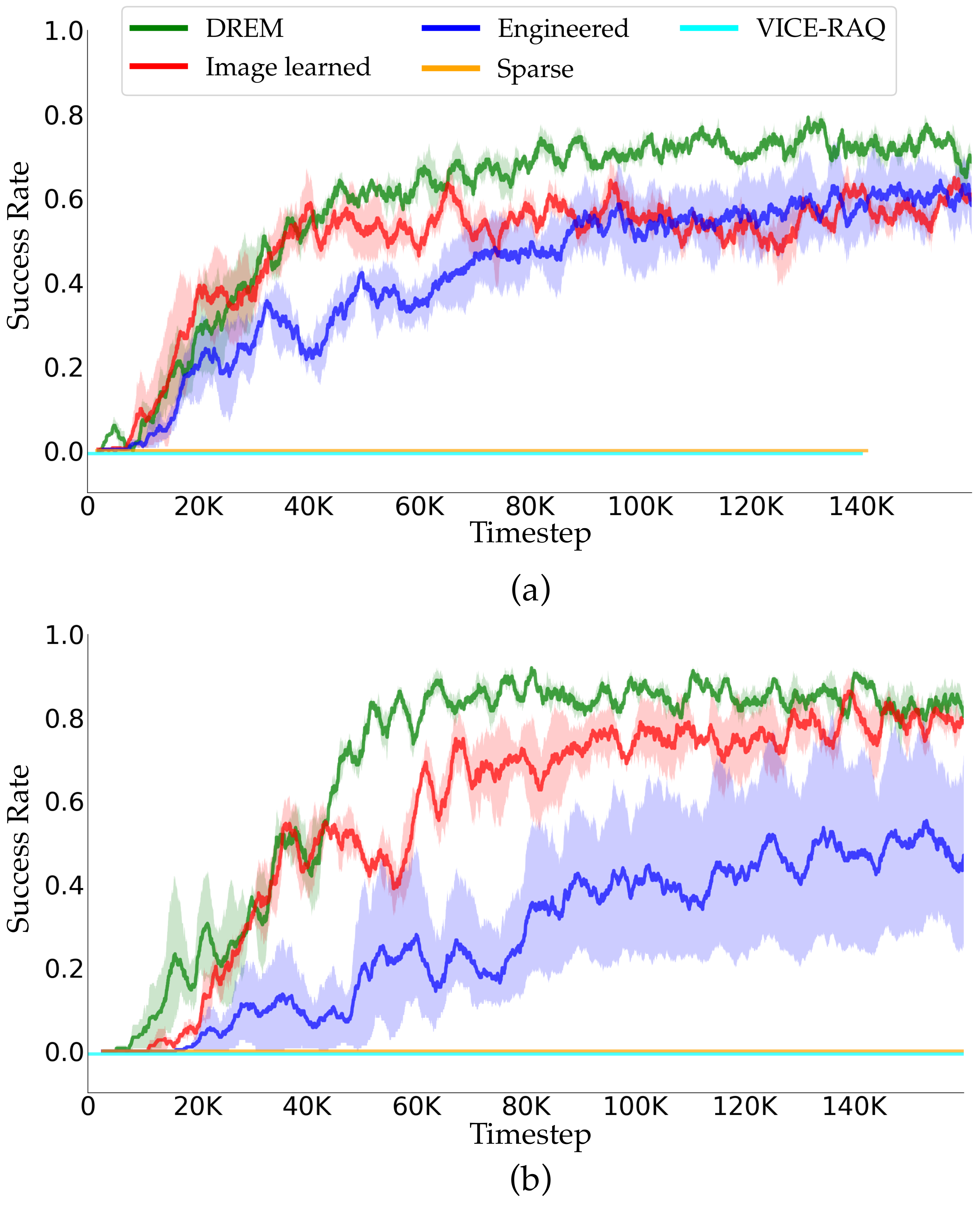}
    \caption{Results of the policy learning experiments on peg-in-hole (a) and USB insertion (b). Each method is run with three different random seeds for each task.}
    \label{fig:experiment:policy_learning}
\vspace{-7mm}
\end{figure}

The policy learning experiments aim to demonstrate the effectiveness of our learned reward, and provide a rigorous comparison with other reward design methods. 
While our approach to reward learning is agnostic to the RL algorithm being used, we use Soft Actor-Critic (SAC) in our evaluations~\cite{haarnoja2018soft}. 
\texttt{DREM} refers to our learned reward model introduced in Section~\ref{sec:method}, with all three modalities as input. We compare it with four baseline methods: \texttt{Sparse} represents the sparse reward, i.e., the binary success indicator at the end of episode; \texttt{Engineered} represents a handcrafted reward $-\exp(\kappa {||x - x_G||_{2})}$, i.e., a function of the $L_2$ distance between the current robot's end-effector pose and the target pose; \texttt{VICE-RAQ} represents the state-of-the-art reward learning algorithm on high-dimensional (image) input~\cite{singh2019end}; \texttt{Image learned} represents our reward model with only images as input. For all baseline experiments, we use RGB images as the input of the policy, while for \texttt{DREM} we use RGB images and force data as the input. We adopted a similar neural network in Section~\ref{sec:method:representation_learning:encoder} as the policy network. The experimental results are shown in Fig.~\ref{fig:experiment:policy_learning}.

It can be observed that our learned reward (\texttt{DREM} and \texttt{Image learned}) outperforms the other reward designs in both peg-in-hole and USB insertion, indicating that our proposed reward learning framework is able to learn reward functions that boosts RL training. The fact that \texttt{DREM} achieves faster convergence and a higher success rate than \texttt{Image learned} demonstrates the advantages of multi-modal input (force-torque signals in our case), i.e., enabling a more efficient and robust policy. \texttt{Sparse} does not achieve any success within the training budget, showing difficulty of the two contact-rich tasks. \texttt{Engineered} performs satisfactorily at peg-in-hole but much worse at the more difficult USB insertion, as it requires a tighter fit and a strict orientation alignment.
It should be noted that \texttt{VICE-RAQ} does not succeed at the two tasks experimented with; the success rate remains 0 after $150$k training steps. We hypothesize that only using images, adversarial training is easily stuck in a local optimal as similar images got similar rewards even though they might be in different contact states. We constructed a significantly easier task by configuring the peg to be close to the hole initially (aligned right above the hole center), to avoid training getting stuck due to frequent contact states. A $0.10$ success rate is achieved after $50$k training steps; however, if the peg is initially misaligned with the hole center, though close, training fails to achieve any success.

\vspace{-1mm}
\section{Related Work}
In this section, we provide a brief review of research on reward learning, with emphasis on applications in contact-rich manipulation. Rewards provide a mechanism for humans to specify tasks to robots \cite{sutton2018reinforcement}. However, reward specification is challenging and can lead to unanticipated behavior \cite{shapiro2002user, amodei2016concrete}. Consequently, approaches to reward learning have attracted increasing attention in the last two decades \cite{chernova2014robot, hadfield2017inverse, scobee2019maximum}, with inverse reinforcement learning (IRL) being the prominent paradigm \cite{argall2009survey}. IRL aims to recover a reward function given expert demonstrations  \cite{atkeson1997robot, ng2000algorithms, abbeel2004apprenticeship}. However, as many rewards may explain the demonstrations equally well, additional objective criteria that help identify the \emph{correct} reward have been explored \cite{bagnell2006maximum, ramachandran2007bayesian, ziebart2008maximum, ho2016generative}. 
For instance, \cite{ziebart2008maximum} utilize the principle of maximum entropy to learn the reward.

Recently, deep variants of IRL have also been developed with the goal of addressing tasks with larger state, observation, and action spaces \cite{finn2016guided, fu2017learning, fu2018variational}. For instance, Finn et al. \cite{finn2016guided} provide guided cost learning, which uses neural networks to parameterize the reward function. \cite{fu2017learning} provide a reward learning approach robust to changes in environment dynamics. These IRL techniques and others have been successfully demonstrated on a diverse set of problems, such as modeling driving behaviors \cite{ziebart2008maximum, kuderer2015learning, wu2020efficient}, robotic manipulation \cite{finn2016guided, muelling2014learning}, and grasping \cite{kalakrishnan2013learning}. Despite the steady success of IRL techniques, several open challenges remain. First, most techniques require hand-engineered features (or states) while efficient learning of rewards from high-dimensional observations remains difficult. Second, state-of-the-art IRL techniques utilize adversarial optimization~\cite{finn2016guided, fu2017learning, fu2018variational}, thereby inheriting the associated training instabilities \cite{salimans2016improved}. Consequently, we seek to develop a reward learning approach that can reason over high-dimensional observations spaces without utilizing adversarial training.

There are a few related attempts along this line of research. Specifically, \cite{wang2019random} re-framed imitation learning within the standard reinforcement learning setting using expert policy support estimation. However, they only evaluated their method on a low-dimensional state space. \cite{cabi2019scaling} proposed to use reward sketching provided by annotators to learn a dense model from the image input. The major drawback of their work is that their method requires many execution trajectories and the corresponding annotations, which are expensive to acquire. By contrast, in our proposed method, the data used to train the reward model is self-generated by the robot. The sampling process only takes a single execution trajectory, much cheaper than~\cite{cabi2019scaling}.

Many research efforts have been devoted in recent years to learning contact-rich manipulation skills, such as peg insertion, block packing, etc. However, most of the works use either images~\cite{levine2016end, sermanet2018time, zhu2018reinforcement} or haptic feedback~\cite{kalakrishnan2011learning, tian2019manipulation, sung2017learning} as the policy input. There are only a few works that exploit image and force together. Specifically, \cite{fazeli2019see} combined vision and force to learn to play Jenga, \cite{lee2019making} proposed to learn a multi-modal representation of sensor inputs and utilize the representation for policy learning on peg insertion tasks, and the authors in  \cite{kappler2015data} achieved motion generation for manipulation with multi-modal sensor feedback using manipulation graphs. However, all the previous works focus on policy learning for manipulation tasks, while our method tries to learn the reward function that can be seamlessly integrated into policy learning algorithms for manipulation. To the best of our knowledge, our work is the first attempt to learn dense rewards from multi-modal inputs.

\vspace{-1mm}
\section{Conclusion and Future Work}
\label{sec:conclusion}
\vspace{-1mm}
In summary, we propose an approach to learn dense rewards from multi-modal observations (\texttt{DREM}), with particular emphasis on contact-rich manipulation tasks. A novel backward sampling method is proposed to generate requisite training data through exploration, guided by a single expert demonstration. Dense rewards are learned from the high-dimensional observation space without adversarial training. We evaluate \texttt{DREM} on two assembly tasks and demonstrate its efficacy through comparisons with other learned and handcrafted reward functions.

Our work also motivates several future directions. For instance, currently \texttt{DREM} is geared towards \emph{monotonic} tasks, i.e., where task progress is monotonic in the observation space. While this assumption is valid for a range of assembly tasks (due to the presence of defined start and goal states), extension of \texttt{DREM} to non-monotonic tasks remains of interest.
Further, the experiments indicate that task-specific logical concepts (such as in-contact or losing-contact) underlie our learned reward. Hence, another interesting direction is to formalize the problem of recovering such logical concepts from dense rewards to enable multi-task learning and consequently further improve sample efficiency.










\bibliographystyle{IEEEtran}
\bibliography{root}

\end{document}